\documentclass[letterpaper]{article} 
\usepackage{aaai2026}  
\usepackage{times}  
\usepackage{helvet}  
\usepackage{courier}  
\usepackage[hyphens]{url}  
\usepackage{graphicx} 
\usepackage{natbib}  
\usepackage{caption} 
\usepackage{algorithm}
\usepackage{algorithmic}

\usepackage{newfloat}
\usepackage{listings}
\DeclareCaptionStyle{ruled}{labelfont=normalfont,labelsep=colon,strut=off} 
\floatstyle{ruled}
\newfloat{listing}{tb}{lst}{}
\floatname{listing}{Listing}
\usepackage{multirow}
\usepackage{makecell}   
\usepackage{amssymb}    
\usepackage{tabularray}
\usepackage{amsmath}
\usepackage{booktabs}
\usepackage{anyfontsize} 

\usepackage[colorinlistoftodos]{todonotes}
\usepackage{soul}  

\title{Refinement Contrastive Learning of Cell-Gene Associations for Unsupervised Cell Type Identification}
\author{
    Liang Peng\textsuperscript{\rm 1}\equalcontrib, Haopeng Liu\textsuperscript{\rm 1}\equalcontrib, Yixuan Ye\textsuperscript{\rm 1}\equalcontrib, Cheng Liu\textsuperscript{\rm 2,1}\thanks{Corresponding author}, Wenjun Shen\textsuperscript{\rm 3}, Si Wu\textsuperscript{\rm 4}, Hau-San Wong\textsuperscript{\rm 5}
}
\affiliations{
    \textsuperscript{\rm 1}Department of Computer Science, Shantou University \\
    \textsuperscript{\rm 2}College of Computer Science and Technology, Huaqiao University \\ 
    \textsuperscript{\rm 3}Shantou University Medical College, Shantou University \\
    \textsuperscript{\rm 4}School of Computer Science and Engineering, South China University of Technology \\
    \textsuperscript{\rm 5}Department of Computer Science, City University of Hong Kong \\
    \{23lpeng, 23hpliu, 22yxye2, wjshen\}@stu.edu.cn, chengliu10@gmail.com, cswusi@scut.edu.cn, cshswong@cityu.edu.hk
}

\usepackage{bibentry}

\begin{document}

\maketitle

\begin{abstract}
Unsupervised cell type identification is crucial for uncovering and characterizing heterogeneous populations in single cell omics studies. Although a range of clustering methods have been developed, most focus exclusively on intrinsic cellular structure and ignore the pivotal role of cell-gene associations, which limits their ability to distinguish closely related cell types. To this end, we propose a Refinement Contrastive Learning framework (\textbf{scRCL}) that explicitly incorporates cell-gene interactions to derive more informative representations. 
Specifically, we introduce two contrastive distribution alignment components that reveal reliable intrinsic cellular structures by effectively exploiting cell-cell structural relationships.
Additionally, we develop a refinement module that integrates gene-correlation structure learning to enhance cell embeddings by capturing underlying cell-gene associations. This module strengthens connections between cells and their associated genes, refining the representation learning to exploiting biologically meaningful relationships.
Extensive experiments on several single‑cell RNA‑seq and spatial transcriptomics benchmark datasets demonstrate that our method consistently outperforms state-of-the-art baselines in cell-type identification accuracy. Moreover, downstream biological analyses confirm that the recovered cell populations exhibit coherent gene‑expression signatures, further validating the biological relevance of our approach. The code is available at https://github.com/THPengL/scRCL.
\end{abstract}


\section{Introduction}
Advances in single-cell technologies have enabled high-resolution characterization of gene expression profiles at the individual cell level.
These technologies enable detailed analysis of cellular heterogeneity and have been widely applied in various fields, including the construction of organismal cell atlases \cite{0oca-2015-droplet-klein,0oca-2017-massively-zheng}, developmental biology \cite{0db-2019-resolving-guo}, and clinical research\cite{0cr-2020-single-guo,0cr-2020-single-wilk}.
Unsupervised cell-type identification is essential for uncovering and characterizing heterogeneous cell populations in single-cell omics analysis. It provides critical insights into a range of biological processes, such as cellular differentiation, lineage tracing, and tumor microenvironment characterization.
\begin{figure}[!t]
    \centering{
        \includegraphics[width=0.47\textwidth]{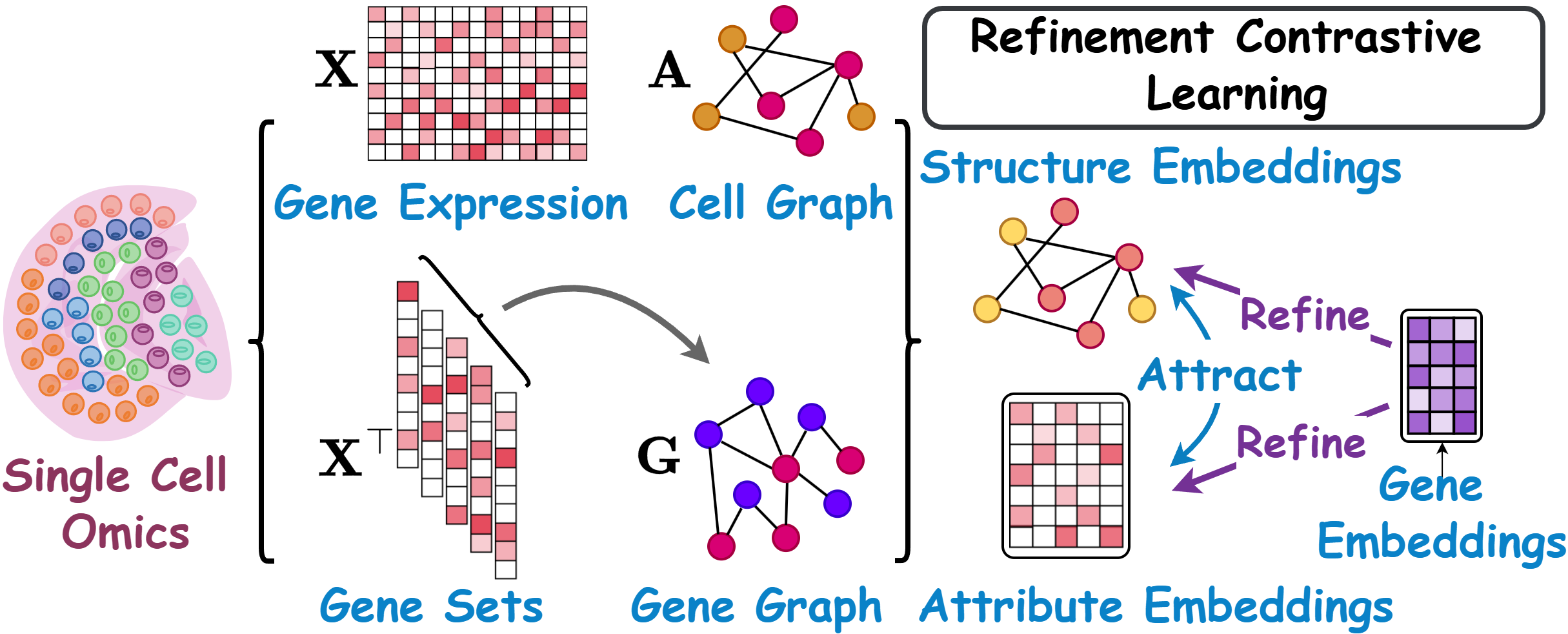}
        \caption{Illustration of our method. Blue lines indicates the contrastive learning component for capturing intrinsic cellular structure using the cell graph. The purple lines represent gene correlation learning module, which refines the contrastive process by modeling cell-gene associations.}
    \label{fig_illustration}
    }
\end{figure}

To this end, various clustering methods have been developed to facilitate cell-type identification \cite{2022scNAME-wan,2025cvvc-ye,2025STAIG-yang}. For example, scDeepCluster \cite{2019scDeepCluster-tian} integrates deep count autoencoder \cite{2019DCA-eraslan} with the deep embedded clustering \cite{2016DEC-xie} for cell-type identification from scRNA-seq data, while scziDesk \cite{2020scziDesk-chen} applies weighted soft K-means to denoise latent features. 
Recently, scDFC \cite{2023scDFC-hu} leverages graph attention (GAT) to incorporate cell-cell relationships, enhancing robustness against noise. 
Moreover, for spatial transcriptomics data, which includes spatial context from tissue, several approaches\cite{2021SpaGCN-hu,2022STAGATE-dong,2025COME-Wei} have been proposed that incorporate structural learning to model spatial relationships. These methods aim to improve the accuracy of cell-type identification by leveraging spatial information inherent to the tissue architecture. For instance, GraphST~\cite{2023GraphST-long} integrates graph neural networks with contrastive learning to learn spatial proximity relationships and discriminative representations for spatial clustering. While STAIG~\cite{2025STAIG-yang} incorporates gene expression, spatial context, and histological information to dynamically adjust the graph structure under the guidance of tissue images, thereby enabling more effective representation learning.

However, most existing cell-type identification methods focus exclusively on capturing intrinsic cellular structure, often overlooking the underlying cell-gene interactions. From a biological perspective, genes interact through regulatory pathways, and distinct cell types are characterized by specific sets of marker genes. These relationships suggest that the intrinsic structures of both cells and genes are tightly coupled, jointly shaping the overall cell-gene associations. Effectively modeling these associations is essential for accurately distinguishing cell types and uncovering their underlying biological functions. 

To address this, we propose a Refinement Contrastive Learning framework (\textbf{scRCL}.) that explicitly incorporates cell-gene interactions to learn more informative and biologically meaningful representations.
Specifically, we introduce a contrastive distribution alignment module that leverages topological tissue information to uncover reliable intrinsic cellular structures, alongside a gene-correlation structure learning module designed to capture gene-gene relationships.
More importantly, as illustrated in Figure \ref{fig_illustration}, we further develop a refinement module that leverages both gene-gene correlation structures and intrinsic cellular structures to effectively model the underlying cell-gene associations. This module reinforces the coupling between cells and their associated genes, thereby refining the learned representations to emphasize biologically relevant patterns.

The main contributions of this study can be summarized as follows: 
1) We develop a novel contrastive distribution alignment module that effectively leverages topological information to uncover reliable intrinsic cellular structures. This module is broadly applicable to various types of single-cell omics data, including scRNA-seq and spatial transcriptomics data. 2) Instead of treating cells and genes independently, the proposed refinement module jointly models their interactions, allowing for the identification of underlying cell-gene associations across coherent cellular populations. 3) Extensive experiments on diverse scRNA-seq and spatial transcriptomics datasets validate the effectiveness of our method. Additionally, a series of biological analyses confirm that the identified cell populations exhibit coherent gene-expression patterns, demonstrating the biological relevance of our approach.
    
    

\section{Related Work}
In this section, we review unsupervised cell-type identification methods developed for both scRNA-seq and spatial transcriptomics data, highlighting their underlying principles and key differences from the proposed approach.

\subsubsection{scRNA-seq Cell Type Identification.}
scRNA-seq enables transcriptome-wide gene expression profiling at single-cell resolution, facilitating the investigation of cellular heterogeneity in complex tissues. A wide range of computational methods have been proposed for unsupervised cell-type identification from scRNA-seq data. Traditional approaches such as PCA-based clustering \cite{2016pcaReduce-vzurauskiene} and CIDR \cite{2017CIDR-lin} operate on dimensionally reduced gene expression profiles or pairwise cell similarity matrices. In addition, deep learning-based methods have gained popularity due to their ability to capture complex, nonlinear relationships within high-dimensional single-cell data and to learn robust low-dimensional representations that facilitate more accurate cell clustering.
For example, DCA \cite{2019DCA-eraslan} models gene expression with a zero-inflated negative binomial (ZINB) distribution for denoising and representation learning. scDeepCluster \cite{2019scDeepCluster-tian} combines DCA with DEC \cite{2016DEC-xie} for joint optimization. scNAME \cite{2022scNAME-wan} and scDCCA \cite{2023scDCCA-wang} adopt contrastive learning to enhance pairwise cell proximity modeling. 
Recently, GNN-based methods \cite{2021scGNN-wang,2021scGAE-luo} have emerged to more effectively capture cell-cell interactions by leveraging the underlying graph structure of single-cell data. For instance, scDSC \cite{2022scDSC-gan} adopts a mutual supervised strategy to unify a ZINB model-based autoencoder and a GNN module to guide the clustering task. While scDFC \cite{2023scDFC-hu} utilizes GATs to adaptively assign weights to neighboring cells, thereby enhancing robustness to noise. 

\begin{figure*}[!t]
    \centering{
        \includegraphics[width=0.93\textwidth]{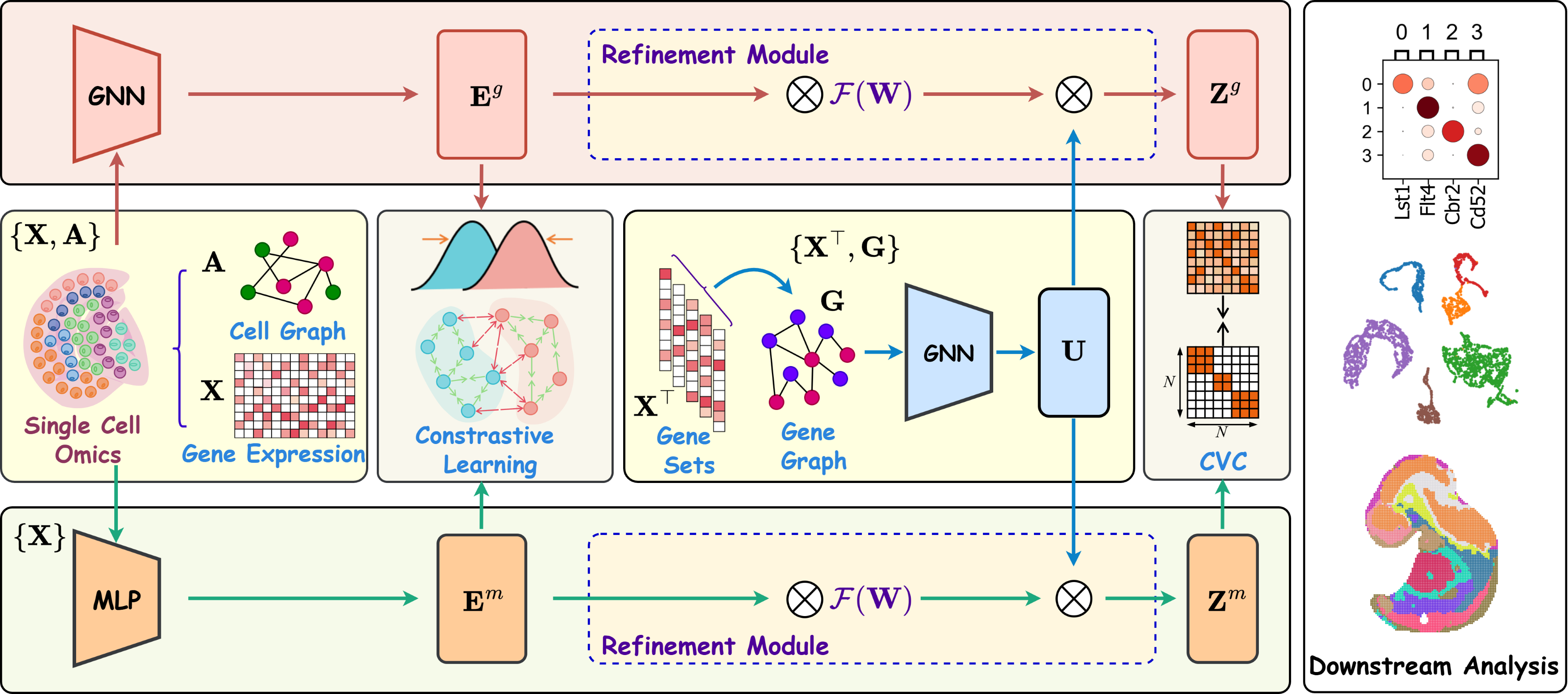}
        \caption{Overview of the proposed scRCL. It extracts cell- and gene-level embeddings from gene expression data and graphs via GNNs and MLPs. A refinement module with contrastive learning aligns these embeddings to obtain informative representations for downstream analyses, including marker gene detection, cell type identification, and spatial domain visualization.}
    \label{fig_overview}
    }
\end{figure*}
\subsubsection{Domain Identification for Spatial Transcriptomics.}
Spatial transcriptomics further extends the capabilities of single-cell analysis by preserving the spatial context of gene expression, offering valuable insights into tissue organization. A key challenge in this domain lies in effectively integrating spatial proximity information with gene expression data to enhance the accuracy of spatial domain identification. To address this, various methods have been proposed \cite{2022STAGATE-dong,2024MAFN-zhu,2024MuCoST-zhang}. For example, SpaGCN \cite{2021SpaGCN-hu} and DeepST \cite{2022DeepST-xu} leverage graph neural networks (GNNs) to jointly model gene expression profiles, spatial coordinates, and even histology images, enabling more accurate and biologically meaningful identification of spatial domains. 
GraphST \cite{2023GraphST-long} incorporates neighborhood contrastive learning to minimize embedding distances among neighboring samples, improving discriminative  representations. 
More recently, MAEST \cite{2025MAEST-zhu} tackles the issue of high dropout rates by employing a graph masked autoencoder, which effectively reconstructs gene expression signals from sparse input. While STAIG \cite{2025STAIG-yang} dynamically adjusts graphs under the guidance of histology images to perform graph contrastive learning.

However, most existing cell identification methods for both scRNA-seq and spatial transcriptomics data overlook the underlying cell-gene interactions, which are crucial for understanding cellular function. Even though scGCOT \cite{2024scGCOT-yu} employs GATs to jointly learn cell-cell and gene-gene correlations for reconstructing the original gene expression matrix, it primarily focuses on expression reconstruction and fails to capture discriminative cellular structures. Moreover, its applicability is limited to scRNA-seq data, without consideration of spatial context.
In contrast, our method explicitly incorporates cell-gene associations. It leverages gene-informed refinement and contrastive learning to enhance cell representation learning, which is generalizable across scRNA-seq and spatial transcriptomics data.

\section{Methodology}
In this section, we present our method for cell-type identification, which leverages cell-gene associations to enhance representation learning. An overview of the overall framework is illustrated in Figure~\ref{fig_overview}.

\subsection{Preliminaries}
Suppose we have a gene expression matrix $\textbf{X} \in \mathbb{R}^{N \times M}$, where $N$ and $M$ are the number of cells and genes, as the input feature. 
For spatial transcriptomics data, a cell graph $\textbf{A} \in \{0, 1\}^{N \times N}$ can be constructed using spatial coordinates via the 
$k$-nearest neighbors (KNN) algorithm. In the case of scRNA-seq data, where spatial information is not available, $\textbf{A}$ can be estimated directly from the expression matrix $\textbf{X}$ using similarity metric.
Meanwhile, to estimate the initial correlation structure among genes, we consider the transposed expression matrix
$\textbf{X}^{\top} \in \mathbb{R}^{M \times N}$ as gene sets, where each row corresponds to a gene's expression profile across all cells. Based on this representation, a gene graph
$\textbf{G} \in \{0, 1\}^{M \times M}$ can be estimated. 

\subsection{Heterogeneous Embedding Distribution Alignment}
To learn reliable cell embeddings, we develop a contrastive learning strategy based on the gene expression matrix $\textbf{X}$ and the cell graph $\textbf{A}$. Instead of relying on predefined data augmentation or graph perturbation techniques, we adopt a heterogeneous embedding approach (as shown in Fig. \ref{fig_modeldesign} left, which empirically outperforms configurations using either two GCNs or two MLPs). Specifically, we utilize graph convolutional networks (GCNs) to encode $\textbf{X}$ in conjunction with $\textbf{A}$, and employ traditional multi-layer perceptrons (MLPs) to encode $\textbf{X}$ independently. This design allows the model to capture both graph-structured and feature-based representations for effective contrastive learning.
In detail, given an MLPs encoder $\phi$ and a GCNs encoder $\psi$, we derive the latent embeddings $\textbf{E}^m$and $\textbf{E}^g$, respectively:
\begin{equation}\label{eq_encoders}
\begin{split}
    &\textbf{E}^{m} = \phi(\textbf{X}),
    ~\textbf{E}^{g} = \psi(\textbf{X}, \textbf{A}).
\end{split}
\end{equation}

To enable effective semantic information sharing between the heterogeneous embeddings $\textbf{E}^m$ and $\textbf{E}^g$, we introduce an embedding distribution alignment strategy that aligns the representations from both a global and a cell-level (local) perspective. Formally, let $p^{global}(\textbf{E}) \in \mathbb{R}^{N \times d}$ denotes the softmax-generated overall embedding distribution across the entire cell population, where $d$ is the latent dimension. 
\begin{equation}\label{p_view}
    p^{global}=\begin{bmatrix}
         p_{11} & ...  & p_{1d} \\
         \vdots  & \ddots  & \vdots \\
         p_{N1} & ... & p_{Nd} 
    \end{bmatrix}.
\end{equation}
Similarly, $p^{cell}(\textbf{E}_i) \in \mathbb{R}^{d}$ denotes the probability distribution for individual cell $i$'s embedding after softmax, capturing its local semantic features.
To achieve bidirectional alignment, we minimize the Symmetric Kullback-Leibler (SKL) divergence\cite{2021SKL-RDrop,2025SKL-NeuCGC-peng,2025smart-Peng} between the respective distributions. Specifically, we align the representations by minimizing:
\begin{equation}\label{eq_L_GDA}
\begin{split}
    \mathcal{L}_{HEA} = &\textrm{SKL}(p^{global}(\textbf{E}^m), p^{global}(\textbf{E}^g)) + \\ &\sum_{i=1}^N \textrm{SKL}(p^{cell}(\textbf{E}_i^m), p^{cell}(\textbf{E}_i^g)),
\end{split}
\end{equation}
where $\text{SKL}(P, Q) = \text{KL}(P||Q) + \text{KL}(Q||P)$ measures the symmetric divergence between two probability distributions $P$ and $Q$ in a bidirectional manner. 
In our framework, this metric is used to reduce the distributional discrepancy between the global embeddings
 $p^{global}(\textbf{E}^m)$ and $p^{global}(\textbf{E}^g)$, as well as between each cell-level embedding pair $p^{cell}(\textbf{E}_i^m)$ and $p^{cell}(\textbf{E}_i^g)$. 
 This strategy encourages the model to extract complementary and consistent semantic information from both the structural view (captured by GCNs) and the attribute view (captured by MLPs). Thus, by aligning feature distributions across these heterogeneous embeddings, the model effectively mitigates inconsistencies arising from their distinct inductive biases.

\subsection{Cellular Neighborhood Distribution Contrastive Alignment}
Beyond global and cell counterpart alignment, cellular graphs encode rich local neighborhood semantics: connected cells may share similar gene expression profiles (functional homology) or spatial proximity (structural contiguity), implying higher likelihood of shared biological identity compared to distant cells. Inspired by graph neighborhood contrastive learning, we introduce a neighborhood distribution contrastive alignment technique to further capture intra-class semantic consistency and enhance inter-class discriminability. Specifically, for each anchor cell, we consider the embedding pair from its two views (i.e., $\textbf{E}_i^m$ and $\textbf{E}_i^g$) and those of its neighbors as positive pairs, while treating embeddings from other cells as negative pairs. To promote semantic coherence within neighborhoods and improve representation separability, we minimize the SKL divergence between embeddings of positive pairs while maximizing that for negative pairs. This leads to the following objective:
\begin{equation}\label{eq_L_NCA}
    \mathcal{L}_{NDC} = \frac{1}{N} \sum_{i=1}^N \frac{\frac{1}{|\mathcal{N}_i|+1} (\kappa_{ii} + \sum_{c_k \in {\mathcal{N}_i}} \kappa_{ik})}
{\frac{1}{N-1}\sum_{j=1, j \ne i} \kappa_{ij}},
\end{equation}
where $\mathcal{N}_i$ is the neighbor set within graph $\textbf{A}$, and $\kappa \in \mathbb{R}^{N \times N}$ is the instance-level pairwise cross-view SKL divergence matrix, defined as:
\begin{equation}\label{eq_cmp_K}
    \kappa_{ij} = \text{SKL}(p^{cell}(\mathbf{E}_i^m), p^{cell}(\mathbf{E}_j^g)).
\end{equation}
Due to the non-negativity of SKL divergences, for arbitrary distinct cells $i, j$, we have $\kappa_{ij} > 0$. 
This neighborhood-aware alignment encourages the model to learn shared semantic signals within structural contexts, while maximizing discriminative margins between unrelated cells. It serves as a complementary supervision signal alongside global and cell counterpart distribution alignment, facilitating more coherent and semantically meaningful embedding spaces.

\subsection{Cell-Gene Interaction-Aware Representation Refinement}
In biological systems, genes interact through regulatory pathways, and distinct cell types are characterized by specific of marker gene sets. Unlike most prior approaches that focus solely on learning representations from gene expression profiles to capture intrinsic cellular properties, our method incorporates a dedicated module for refining cell embeddings by leveraging cell-gene associations in addition to feature distribution alignment. Specifically, we introduce a gene-informed refinement module that leverages gene-level information structured by gene-gene co-expression. To obtain informative gene representations, we employ a GCN-based gene encoder $\varphi(\cdot)$ that takes  transposed gene expression matrix (gene sets) $\textbf{X}^\top$ and gene graph $\textbf{G}$ as inputs: 
\begin{equation}\label{eq_encoders}
    \textbf{U} = \varphi(\textbf{X}^{\top}, \textbf{G}),
\end{equation}
where $\textbf{U}$ denotes the latent embedding matrix of all genes, capturing the underlying gene-gene interaction structure.
To explicitly encode intrinsic cell-gene associations informed by this structure, we further leverage the gene embeddings $\textbf{U}$ to refine the learning of cell representations:
\begin{equation}\label{eq_encoders}
\begin{split}
    \textbf{Z}^m &= \textbf{E}^m \mathcal{F} (\mathbf{W}) \textbf{U}, \\
    \textbf{Z}^g &= \textbf{E}^g \mathcal{F} (\mathbf{W}) \textbf{U}.
\end{split}
\end{equation}
where $\textbf{Z}^m,\textbf{Z}^g \in \mathbb{R}^{N \times c}$, $c$ is the number of cell types, and $\mathcal{F}(\textbf{W})$ denotes a refinement function that bridges the cell embeddings $\textbf{E}$ and $\textbf{U}$. In our implementation, we simplify $\mathcal{F}$ to a linear transformation parameterized by the learnable matrix $\textbf{W}$. This can be viewed as an inverse tri-matrix factorization framework, which explicitly models the structural interactions between cells and genes to enhance representation learning. In practice, this enables each cell to integrate information from its most relevant genes, e.g., highly expressed or cell-type-specific marker genes, thus improving the biological coherence of the learned cell embeddings.

\begin{table*}[!ht]
    \centering
    \setlength{\tabcolsep}{2.2mm}{
    {\fontsize{7.5pt}{7.5pt}\selectfont 
    \begin{tabular}{l| ccc| ccc| ccc| ccc| ccc}
    \toprule 
    \multirow{2}*{\textbf{Method}} & \multicolumn{3}{c|}{\textbf{Tumor}} & \multicolumn{3}{c|}{\textbf{Diaphragm}} & \multicolumn{3}{c|}{\textbf{Lung}} & \multicolumn{3}{c|}{\textbf{Trachea}} & \multicolumn{3}{c}{\textbf{Human\_ESC}} \\
    &\textbf{ACC} &\textbf{NMI}  &\textbf{ARI} &\textbf{ACC} &\textbf{NMI} &\textbf{ARI} &\textbf{ACC} &\textbf{NMI} &\textbf{ARI} &\textbf{ACC} &\textbf{NMI} &\textbf{ARI} &\textbf{ACC} &\textbf{NMI} &\textbf{ARI} \\
    \midrule
    \textbf{scDeepCluster}
     &35.76 &29.34 &11.25 &80.21 &83.15 &68.57 &54.79 &73.35 &37.47 &40.21 &57.78 &22.76 &\underline{91.34} &\underline{95.77} &\underline{91.07} \\
    \textbf{scziDesk}
     &\underline{78.07} &41.76 &43.00 &98.02 &92.41 &94.88 &73.20 &78.66 &76.64 &82.24 &63.42 &64.90 &75.74 &89.80 &72.77 \\ 
     \textbf{GraphSCC}
     &61.38 &\underline{62.58} &38.94 &99.01 &\underline{96.31} &\underline{98.45} &59.22 &74.64 &54.35 &81.67 &69.15 &65.82 &82.36 &92.06 &77.18 \\
    \textbf{scGAE}
     &42.90 &40.91 &35.23 &57.40 &59.55 &31.94 &57.40 &57.59 &35.45 &57.30 &47.51 &31.01 &55.10 &55.98 &33.76 \\
    \textbf{scGAC}
     &68.98 &44.11 &\underline{43.80} &98.05 &93.04 &96.06 &51.37 &71.43 &45.86 &69.85 &68.03 &56.85 &82.53 &91.66 &77.00 \\ 
     \textbf{scNAME}
     &57.08 &48.13 &29.25 &98.69 &94.41 &96.89 &64.88 &78.56 &62.41 &81.60 &67.48 &62.57 &75.89 &89.74 &72.75 \\
    \textbf{scDFC}
     &45.62 &32.02 &16.30 &98.21 &93.49 &96.46 &44.36 &65.56 &32.55 &\underline{95.88} &\underline{86.59} &\underline{88.81} &82.87 &91.91 &77.33 \\
     \textbf{scDCCA}
     &47.88 &43.31 &23.67 &90.51 &86.95 &86.67 &64.70 &75.57 &67.90 &70.98 &59.64 &55.35 &84.09 &92.91 &78.42 \\
    \textbf{SCEA}
     &65.90 &42.24 &36.13 &98.16 &93.08 &96.14 &59.45 &75.75 &54.33 &69.81 &58.15 &52.82 &82.55 &91.33 &77.38 \\ 
     \textbf{AttentionAE-sc}
     &43.40 &48.03 &22.26 &92.10 &88.71 &89.63 &\underline{78.10} &81.41 &\underline{82.09} &79.60 &73.50 &67.78 &83.90 &89.66 &72.61 \\
    \textbf{scGAD}
     &33.33 &30.55 &8.01 &98.53 &94.29 &96.95 &53.63 &72.39 &44.84 &71.69 &62.89 &51.24 &78.70 &90.01 &74.23 \\
     \textbf{scMAE}
     &60.11 &58.49 &34.47 &\underline{99.13} &96.03 &97.68 &72.22 &\underline{82.19} &73.36 &83.70 &71.37 &65.88 &82.79 &91.81 &77.23  \\
     \textbf{scGCOT}
     & 63.86 & 52.19 & 31.42 
     & 98.62 & 94.09 & 96.47 
     & 66.91 & 77.14 & 70.02 
     & 80.40 & 68.52 & 62.21 
     & 83.89 & 92.43 & 78.14 \\
     \midrule
    \textbf{scRCL} 
     & \textbf{79.67} & \textbf{68.29} & \textbf{59.83} 
     & \textbf{99.63} & \textbf{98.22} & \textbf{99.12} 
     & \textbf{86.47} & \textbf{85.43} & \textbf{89.87} 
     & \textbf{98.00} & \textbf{91.85} & \textbf{94.83} 
     & \textbf{99.69} & \textbf{99.04} & \textbf{99.22} \\
     
    \midrule[1.0pt]
    \multirow{2}*{\textbf{Method}} 
    &\multicolumn{3}{c|}{\textbf{Zeisel}} 
    &\multicolumn{3}{c|}{\textbf{Bladder}} 
    &\multicolumn{3}{c|}{\textbf{Limb\_Muscle}} 
    &\multicolumn{3}{c|}{\textbf{Spleen}} 
    &\multicolumn{3}{c}{\textbf{Baron\_Human}} \\
    &\textbf{ACC} &\textbf{NMI} &\textbf{ARI} 
    &\textbf{ACC} &\textbf{NMI} &\textbf{ARI}  
    &\textbf{ACC} &\textbf{NMI} &\textbf{ARI} 
    &\textbf{ACC} &\textbf{NMI} &\textbf{ARI} 
    &\textbf{ACC} &\textbf{NMI} &\textbf{ARI} \\
    \midrule
    \textbf{scDeepCluster} 
     &71.40 &73.19 &61.80 &44.34 &58.14 &33.98 &64.56 &78.91 &52.86 &34.91 &45.42 &16.68 &53.93 &71.35 &42.16 \\
    \textbf{scziDesk}
     &77.30 &65.62 &63.28 &86.63 &84.74 &83.16 &93.17 &88.21 &89.26 &76.10 &42.01 &30.71 &53.11 &64.06 &37.77 \\ 
     \textbf{GraphSCC}
     &81.94 &72.06 &73.82 &76.98 &81.08 &75.85 &80.78 &87.77 &79.45 &91.65 &76.27 &87.99 &65.91 &79.83 &62.47 \\
    \textbf{scGAE}
     &47.50 &49.61 &32.39 &47.60 &45.48 &36.18 &52.50 &54.03 &30.51 &43.70 &35.31 &30.93 &42.40 &42.42 &34.54 \\
    \textbf{scGAC}
     &70.16 &70.31 &66.24 &79.69 &79.50 &76.56 &97.63 &95.00 &97.17 &84.87 &64.42 &77.95 &50.37 &66.23 &43.01 \\ 
     \textbf{scNAME}
     &82.40 &\textbf{73.59} &\underline{75.52} &90.89 &90.91 &89.76 &94.02 &93.35 &93.00 &85.14 &73.61 &70.95 &67.98 &80.41 &67.52 \\
    \textbf{scDFC}
     &63.13 &56.12 &44.26 &\underline{95.58} &\underline{92.67} &\underline{94.20} &81.30 &82.60 &73.61 &82.82 &54.57 &60.46 &58.20 &63.99 &42.80 \\
     \textbf{scDCCA}
     &81.94 &72.06 &73.82 &80.25 &76.33 &74.45 &87.30 &89.27 &88.75 &92.73 &77.41 &88.06 &\underline{73.08} &74.20 &67.31 \\
    \textbf{SCEA}
     &72.55 &68.20 &67.72 &81.87 &78.11 &76.32 &97.05 &93.80 &96.37 &80.69 &55.02 &70.09 &54.96 &67.83 &48.12 \\ 
     \textbf{AttentionAE-sc}
     &80.60 &70.24 &67.41 &86.40 &85.65 &83.45 &96.90 &94.82 &\underline{97.67} &77.00 &55.94 &49.96 &61.50 &66.52 &45.92 \\
    \textbf{scGAD}
     &68.03 &64.99 &56.86 &80.98 &77.70 &75.31 &81.89 &87.02 &79.88 &85.87 &66.83 &71.23 &51.18 &67.01 &43.97 \\
     \textbf{scMAE}
     &\underline{83.50} &71.77 &71.69 &77.29 &76.39 &72.66 &92.62 &93.48 &90.81 &92.23 &\underline{82.44} &84.90 &72.97 &\underline{80.47} &\underline{75.92}  \\
     \textbf{scGCOT}
     &76.79 &62.16 &62.18 &80.76 &80.83 &78.48 &\underline{98.48} &\underline{95.06} &97.44 &\underline{94.98} &80.58 &\underline{89.70} &60.74 &63.52 &42.33 \\  
     \midrule
    \textbf{scRCL} 
     &\textbf{83.53} &\underline{73.46} &\textbf{78.29} &\textbf{99.68} &\textbf{97.92} &\textbf{99.25} &\textbf{99.41} &\textbf{97.80} &\textbf{99.04} &\textbf{97.08} &\textbf{86.89} &\textbf{93.36} &\textbf{90.02} &\textbf{87.09} &\textbf{90.69} \\
    \bottomrule
    \end{tabular}
    }
    }
    \captionsetup{justification=raggedright, singlelinecheck=false}
    \caption{Comparison of average clustering performance under five runs on 10 real scRNA-seq datasets.}
    \label{tb_Comparison}
\end{table*}
In addition, we further enhance the consistency and structural fidelity of the learned embeddings through a cross-view correlation contrastive learning mechanism.
Specifically, we define $\textbf{S}_{ij}$ as the cross-view cosine similarity between cells $i$ and $j$, based on their embeddings from two distinct refined views, $\textbf{Z}^{m}$ and $\textbf{Z}^{g}$. The similarity is computed as: 
\begin{equation}\label{eq_cosine_sim}
    \textbf{S}_{ij}=\frac{\textbf{Z}^{m}_i (\textbf{Z}^{g}_j)^{\top}}{||\textbf{Z}^{m}_i||_2||\textbf{Z}^{g}_j||_2}.
\end{equation}
To align the cross-view similarity structure with the underlying cellular topology, we impose a reconstruction loss that encourages the similarity matrix to approximate the adjacency structure (augmented with self-connections):
\begin{equation}\label{eq_cosine_sim}
    \mathcal{L}_{CVC}=\frac{1}{N}||\textbf{S} - (\textbf{A} + \textbf{I})||_F^2.
\end{equation}
This objective enforces structural consistency between embedding views while preserving local connectivity in the cell graph.
Thus, the overall objective function for jointly optimizing all modules is formulated as:
\begin{equation}\label{eq_overall_loss}
\begin{split}
    \mathop{\min}\limits_{\phi, \psi, \varphi, \textbf{W}}~\mathcal{L} = &~~ \mathop{\min}\limits_{\phi, \psi, \varphi, \textbf{W}}~ \mathcal{L}_{HEA} + \alpha \mathcal{L}_{NDC} + 
     \beta \mathcal{L}_{CVC},
\end{split}
\end{equation}
where the coefficients $\alpha,\beta$ balance the loss terms. The encoders $\phi$ and $\psi$ are jointly optimized with respect to all three losses, while the gene encoder $\varphi$ and the trainable parameters $\mathbf{W}$ in the refinement module are updated solely based on $\mathcal{L}_{CVC}$.
By this way, scRCL effectively captures intrinsic cellular structures, gene-gene relationships, and their underlying associations. 
After training, the final representation $\mathbf{Z}$ for each cell is obtained by concatenating the view-specific embeddings:
\begin{equation}\label{eq_overall_loss}
    \textbf{Z} = [\mathbf{Z}^m|\mathbf{Z}^g].
\end{equation}
The resulting embedding $\mathbf{Z}$ is then used for $k$-means clustering to identify distinct cell populations, and serves as a basis for a variety of downstream analyses.



\section{Experiment}
\subsection{Experimental Setup}
\noindent \textbf{Datasets:}
We conduct cell-type identification experiments on a diverse collection of benchmark datasets, encompassing both scRNA-seq datasets: Tumor \cite{0Data-2017Li_Tumor-li}, Diaphragm, Lung, Trachea, Bladder, Limb\_Muscle, Spleen \cite{0Data-2018Quake-schaum}, Human\_ESC \cite{0Data-2016Human_ESC-chu}, Zeisel \cite{0Data-2015Zeisel-zeisel}, Baron\_Human \cite{0Data-2016Baron_Human-baron}, and spatial transcriptomics datasets: LIBD human dorsolateral prefrontal cortex (DLPFC) \cite{0Data-2021DLPFC-maynard} with 12 tissue slices, Stereo-seq E9.5 Mouse Embryo \cite{0Data-2022Mouse_Embryo-chen}, Human Breast Cancer \cite{0Data-2011Human_Breast_Cancer-buache}, and Mouse Brain Anterior \cite{0Data-2007Mouse_Brain-lein}. 

\noindent \textbf{Comparison Approaches:}
We compare our scRCL with several state-of-the-art clustering approaches designed for scRNA-seq and spatial transcriptomics data analysis.
For scRNA-seq, we include: scGCOT \cite{2024scGCOT-yu}, scMAE \cite{2024scMAE-fang}, scGAD \cite{2023scGAD-zhai}, AttentionAE-sc \cite{2023AttentionAE-sc-li}, SCEA \cite{2023SCEA-abadi}, scDCCA \cite{2023scDCCA-wang}, scDFC \cite{2023scDFC-hu}, scNAME \cite{2022scNAME-wan}, scGAC \cite{2022scGAC-cheng}, scGAE \cite{2021scGAE-luo}, GraphSCC \cite{2020GraphSCC-zeng}, scziDesk \cite{2020scziDesk-chen}, and scDeepCluster \cite{2019scDeepCluster-tian}.
For spatial transcriptomics, we compare against: STAIG \cite{2025STAIG-yang}, MAEST \cite{2025MAEST-zhu}, MuCoST\cite{2024MuCoST-zhang}, GraphST \cite{2023GraphST-long}, and Spatial-MGCN \cite{2023spatial-MGCN-wang}.

\noindent \textbf{Evaluation Metrics:}
We assess results using three widely adopted metrics: clustering accuracy (ACC), normalized mutual information (NMI), and adjusted rand index (ARI). Higher values indicate better performance.
The datasets and implementation details are in the Supplementary File.

\begin{figure*}[!t]
    \centering
    \includegraphics[width=0.95\textwidth]{Figs/fig_DLPFC.pdf}
    \captionsetup{justification=raggedright, singlelinecheck=false}
    \caption{The recognized spatial domains in human DLPFC dataset. (A) Boxplots showing ACC, NMI, ARI scores for the six methods applied to the 12 DLPFC slices. The center line in each boxplot represents the median, the box limits indicate the upper and lower quartiles. (B) Clustering result visualizations for slice $\#$151507 of the DLPFC dataset.}
    \label{fig_DLPFC}
\end{figure*}
\begin{table}[!t]
    \centering
    \setlength{\tabcolsep}{1.0mm}{
    {\fontsize{8pt}{8.5pt}\selectfont 
    \begin{tabular}{c| ccc| ccc | ccc}
    \toprule 
    \multirow{2}*{\textbf{Method}} 
    & \multicolumn{3}{c|}{\textbf{ME E9.5}} 
    & \multicolumn{3}{c|}{\textbf{MBA}} 
    & \multicolumn{3}{c}{\textbf{HBC}} \\
     &\textbf{ACC} &\textbf{NMI} &\textbf{ARI} 
     &\textbf{ACC} &\textbf{NMI} &\textbf{ARI} 
     &\textbf{ACC} &\textbf{NMI} &\textbf{ARI} \\
    \midrule
    \textbf{SMGCN} &\underline{52.0} &\underline{52.6} &\underline{38.7} &\underline{47.4} &68.3 &\underline{43.2} &55.4 &55.9 &54.6 \\
    \textbf{GraphST} &43.1 &52.1 &28.7 &42.9 &69.1 &36.5 &56.0 &66.4 &54.3 \\ 
    \textbf{MuCoST} &33.7 &38.4 &13.1 &45.3 &\underline{71.7} &38.8 &47.3 &64.4 &37.7 \\
    \textbf{MAEST} &48.4 &50.3 &30.6 &43.7 &67.5 &36.5 &54.6 &61.0 &51.3 \\
    \textbf{STAIG} &47.0 &52.0 &30.5 &41.9 &71.0 &31.9 &\underline{58.3} &\textbf{69.7} &\underline{56.7} \\
    \midrule
    \textbf{scRCL} &\textbf{58.8} &\textbf{55.4} &\textbf{42.4} &\textbf{51.9} &\textbf{72.3} &\textbf{50.7} &\textbf{65.9}  &\underline{68.2} &\textbf{65.8} \\
    \bottomrule
    \end{tabular}
    }
    }
    \captionsetup{justification=raggedright, singlelinecheck=false}
    \caption{Clustering results on Mouse Embryo E9.5 (ME E9.5), Mouse Brain Anterior (MBA), and Human Breast Cancer (HBC) datasets.}
    \label{tb_ST}
\end{table}

\subsection{Quantitative Results}
Table \ref{tb_Comparison}, Table \ref{tb_ST}, and Figure \ref{fig_DLPFC} present the quantitative and visualized results for cell-type identification on both scRNA-seq and spatial transcriptomics datasets (for DLPFC, which includes 12 tissue slices, we report the averaged results). From these results, we observe the following:
1) Our method consistently outperforms all competing approaches on both types of single-cell omics data. While scGCOT also considers cell-gene relationships for reconstructing gene expression, it may overlook the discriminative intrinsic cellular structure. In contrast, our method explicitly captures both cellular structures and cell-gene associations, leading to significantly better performance.
2) For spatial transcriptomics, scRCL achieves the best performance across all datasets. Furthermore, from the visualization of cell-type identification results within tissue sections (Figure \ref{fig_DLPFC}B), our method reveals more distinct and spatially coherent structures compared to other baseline methods.

Overall, the results demonstrate that scRCL effectively uncovers spatially coherent and biologically meaningful cell populations.

\begin{table}[!t]
    \centering
    \setlength{\tabcolsep}{0.9mm}{
    {\fontsize{8pt}{8.5pt}\selectfont 
    \begin{tabular}{l| ccc| ccc| ccc}
    \toprule 
    \multirow{2}*{\textbf{Method}} 
    & \multicolumn{3}{c|}{\textbf{Trachea}} 
    & \multicolumn{3}{c|}{\textbf{Zeisel}} 
    & \multicolumn{3}{c}{\textbf{Limb\_Muscle}}  \\
     &\textbf{ACC} &\textbf{NMI}  &\textbf{ARI} 
     &\textbf{ACC} &\textbf{NMI}  &\textbf{ARI} 
     &\textbf{ACC} &\textbf{NMI} &\textbf{ARI} \\
    \midrule
    \textit{w/o} $\mathcal{L}_{HEA}$ 
     &96.9 &89.7 &93.2 
     &81.2 &71.9 &76.6
     &98.3 &96.6 &97.4  \\
    \textit{w/o} $\mathcal{L}_{NDC}$ 
     &79.8 &54.4 &64.0 &67.1 &60.4 &56.9
     &90.6 &86.3 &86.5  \\ 
     \textit{w/o} $\mathcal{L}_{CVC}$ 
     &96.2 &87.2 &89.8 
     &73.8 &66.3 &66.4
     &89.3 &90.4 &90.4  \\
     \midrule
     \textbf{scRCL} 
     &\textbf{98.0} &\textbf{91.9} &\textbf{94.8} 
     & \textbf{83.5} & \textbf{73.5} & \textbf{78.3}
     &\textbf{99.4} &\textbf{97.8} &\textbf{99.0} \\
    \bottomrule
    \end{tabular}
    }
    }
    \captionsetup{justification=raggedright, singlelinecheck=false}
    \caption{Ablation study on each objective.}
    \label{tb_Ablation}
\end{table}
\subsection{Ablation Study}
\subsubsection{Model Structure Design.}
\begin{figure}[!t]
    \centering
    \includegraphics[width=0.48\textwidth]{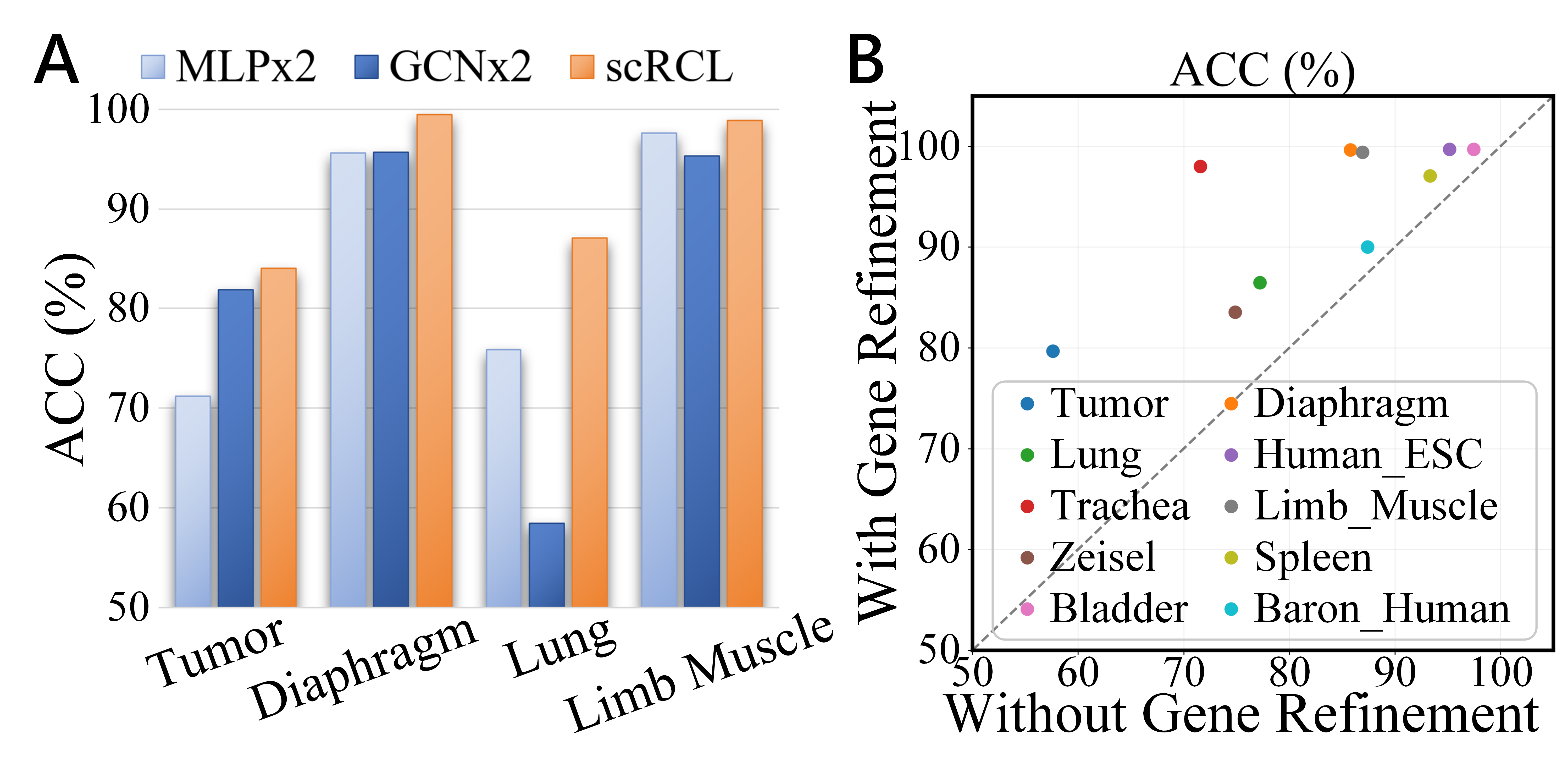}
    \captionsetup{justification=raggedright, singlelinecheck=false}
    \caption{Ablation studies on model structure design.}
    \label{fig_modeldesign}
\end{figure}

We conduct an ablation study on the model structure, focusing on the impact of heterogeneous embedding learning and the refinement module, as illustrated in Figure \ref{fig_modeldesign}. The results show that the heterogeneous embedding strategy outperforms variants that use either two identical GCNs or two identical MLPs encoders, highlighting its effectiveness in capturing complementary structural and attribute information.
Furthermore, the incorporation of the refinement module leads to a significant improvement in cell-type identification performance, highlighting the importance of modeling cell-gene associations for biologically meaningful representation learning.

\begin{figure*}[t]
    \centering
    \includegraphics[width=1.0\textwidth]{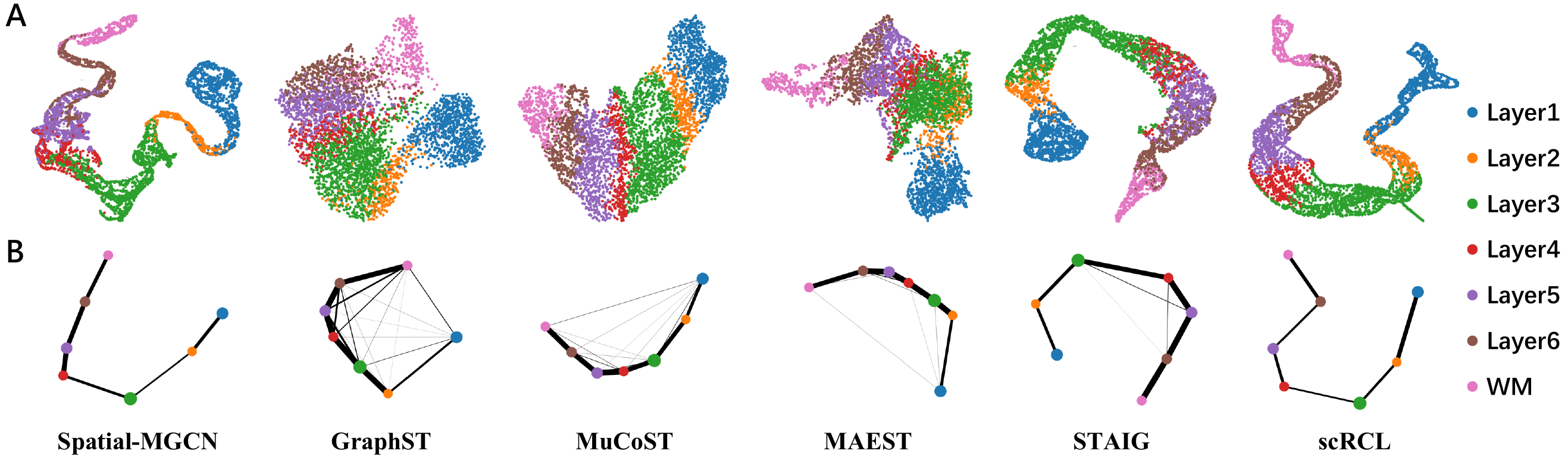}
    \captionsetup{justification=raggedright, singlelinecheck=false}
    \caption{UMAP visualizations (A) and PAGA trajectory inference (B) results derived from the representations of six methods on slice $\#$151507 of the DLPFC dataset.}
    \label{fig_umap_paga}
\end{figure*}
\begin{figure}[!t]
    \centering
    \includegraphics[width=0.48\textwidth]{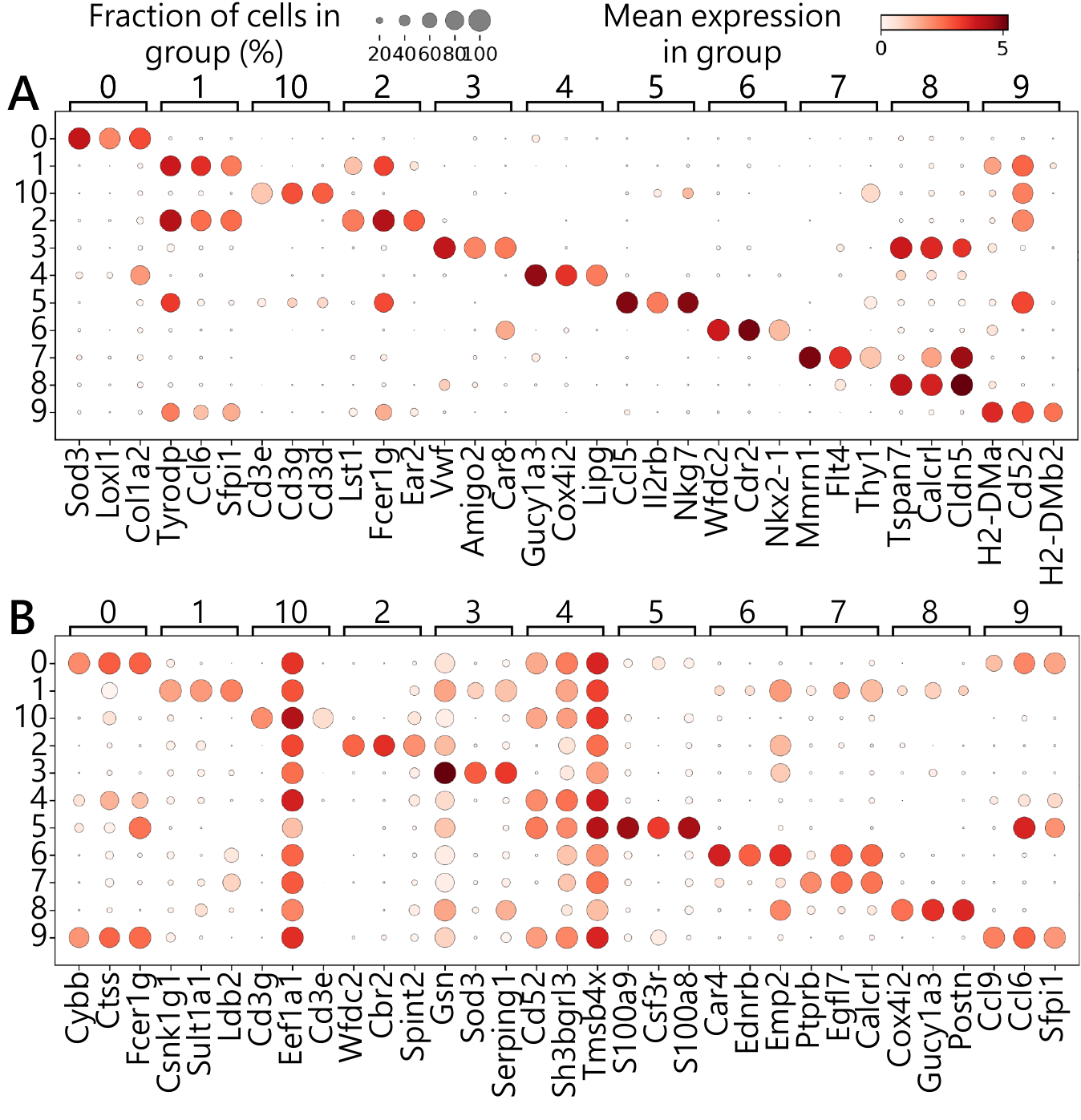}
    \captionsetup{justification=raggedright, singlelinecheck=false}
    \caption{Expression dot-plots of the top three marker genes for each predicted cell label obtained by scRCL with (A) and without (B) the gene-informed refinement module.}
    \label{fig_DEGs}
\end{figure}
\subsubsection{The Impact of Contrastive Learning.}
We present comprehensive ablation studies to evaluate the contribution of each contrastive learning loss in our proposed method ($\mathcal{L}_{HEA}$, $\mathcal{L}_{NDC}$, $\mathcal{L}_{CVC}$). As shown in Table \ref{tb_Ablation}, removing any key loss term associated with the respective module, leads to a noticeable degradation in performance. 
These results highlight the importance of each component in enhancing the overall effectiveness of cell-type identification. 

Overall, the contrastive objectives and model components are seamlessly integrated, working synergistically to capture cellular structures and biologically meaningful cell-gene associations.

\subsection{Biological Analysis}
\subsubsection{Identification of Biologically Meaningful Cell-Type Marker Genes.}
We evaluate the biological interpretability of our predicted clusters by conducting differentially expressed genes (DEGs) analysis to identify marker genes both with and without the refinement module.
For each cluster on the Lung dataset, we compute differential expression scores and select the top three marker genes. Figure \ref{fig_DEGs} presents dot-plot visualizations of these markers across clusters. As illustrated, clusters derived using our full framework (A) exhibit distinctive, highly specific marker-gene expression patterns, whereas clusters obtained without the refinement module (B) show less clear separation. 

We find that the majority of the top marker genes identified by our method align well with known cell-type-specific markers in the CellMarker database \cite{2019CellMarker-zhang}. For example, in the case of cluster 10, the marker genes Cd3e, Cd3d, and Cd3g are highly expressed, which are well-established markers of T cells, enabling accurate annotation of this cluster as T cells. In contrast, without the gene-informed refinement module, cluster 10 is misclassified, with Eef1a1 emerging as a marker gene. However, Eef1a1 is a housekeeping gene and not specific to T cells, indicating reduced specificity and biological interpretability in the absence of refinement.

These results demonstrate that incorporating cell-gene associations not only improves clustering accuracy but also yields biologically meaningful cell-type definitions.

\subsubsection{Trajectory Inference.}
For spatial transcriptomics data, the learned embeddings from scRCL reveal clear spatial differentiation trajectories. Figure~\ref{fig_umap_paga} presents the UMAP \cite{2018UMAP-mcinnes} visualizations of the learned embeddings and corresponding trajectory inference results obtained using PAGA \cite{2019paga-wolf}, comparing scRCL with several baseline approaches. Notably, our method produces well-separated clusters for each cortical layer and the corresponding PAGA graph infers a coherent linear trajectory progressing from layer 1 to layer 6 and then to the white matter. This reflects a biologically consistent spatial organization, where adjacent layers exhibit smooth transitions.
Although Spatial-MGCN is capable of recovering a similar linear trajectory, it fails to clearly separate layers 3, 4 and 5 in the embedding space. Other competing methods display varying degrees of layer overlap, with their inferred trajectories deviating from the expected linear progression. These results demonstrate the superiority of our approach in capturing both spatial structure and underlying developmental trajectories.

\section{Conclusion}
In this work, we proposed a novel unsupervised framework scRCL for cell-type identification that explicitly incorporates cell-gene associations to enhance representation learning from single-cell omics data. Our method effectively captures intrinsic cellular structures, gene-gene interactions, and their underlying associations. It provides a generalizable solution that bridges gene expression profiles with structural cues derived from cell-gene associations, which is applicable to both scRNA-seq and spatial transcriptomics data. This framework offers new insights for downstream analysis in single-cell biology.

\section*{Acknowledgments}
This work was supported in part by Scientific Research Innovation Capability Support Project for Young Faculty (Grant No. ZYGXQNJSKYCXNLZCXM-H8), in part by Guangdong Basic and Applied Basic Research Foundation (Grant No. 2025A1515011692, 2023A1515030154), in part by National Natural Science Foundation of China (Project No.62106136, No. 62072189), in part by TCL Science
and Technology Innovation Fund (Grant No. 20231752), in
part by the Research Grants Council of the Hong Kong Special Administration Region (Grant No. CityU 11206622).

\bibliography{aaai2026}

\end{document}